\documentclass[letterpaper, 10 pt, journal, twoside]{IEEEtran}
\usepackage{subfig,float}
\usepackage{cite}
\usepackage{amsmath,amssymb,amsfonts}
\usepackage{algorithmic}
\usepackage{graphicx}
\usepackage{booktabs}

\usepackage{textcomp}
\usepackage{xcolor}
\def\BibTeX{{\rm B\kern-.05em{\sc i\kern-.025em b}\kern-.08em
    T\kern-.1667em\lower.7ex\hbox{E}\kern-.125emX}}

\usepackage{upgreek}
\usepackage{mathtools}
\usepackage{stmaryrd}
\usepackage{algorithm}
\usepackage{comment}

\newcommand{\R}{\mathbb{R}}

\newcommand{\EX}{\mathfrak{X}}
\newcommand{\ex}{\mathfrak{x}}

\newtheorem{problem}{Problem}
\newtheorem{remark}{Remark}
\newtheorem{assumption}{Assumption}

\begin{document}
\title{Robust Embodied Self-Identification of Morphology in Damaged Multi-Legged Robots}

% \author{Michael~Shell,~\IEEEmembership{Member,~IEEE,}
%         John~Doe,~\IEEEmembership{Fellow,~OSA,}
%         and~Jane~Doe,~\IEEEmembership{Life~Fellow,~IEEE}% <-this % stops a space
% \thanks{M. Shell was with the Department
% of Electrical and Computer Engineering, Georgia Institute of Technology, Atlanta,
% GA, 30332 USA e-mail: (see http://www.michaelshell.org/contact.html).}% <-this % stops a space
% \thanks{J. Doe and J. Doe are with Anonymous University.}% <-this % stops a space
% \thanks{Manuscript received April 19, 2005; revised August 26, 2015.}}
\author{Sahand Farghdani, Mili Patel and Robin Chhabra, ~\IEEEmembership{Senior Member,~IEEE}%
\thanks{S. Farghdani and Mili Patel are with the Mechanical and Aerospace Engineering
Department, Carleton University, Ottawa, ON K1S 5B6, Canada.{\tt\small sahandfarghdani@cmail.carleton.ca, 
        milipatel@cmail.carleton.ca}%\\
        \\
\indent Robin Chhabra (corresponding author) is with the Mechanical, Industrial, and Mechatronics Engineering
Department, Toronto Metropolitan University, Toronto, ON M5B 2K3, Canada.
{\tt\small robin.chhabra@torontomu.ca}}}%


\markboth{IEEE Robotics and Automation Letters.}
% {Farghdani \MakeLowercase{\textit{et al.}}: 
{Damage identification in multi-legged robots} 




\maketitle

\begin{abstract}

Multi-legged robots (MLRs) are vulnerable to leg damage during complex missions, which can impair their performance. This paper presents a self-modeling and damage identification algorithm that enables autonomous adaptation to partial or complete leg loss using only data from a low-cost IMU. A novel FFT-based filter is introduced to address time-inconsistent signals, improving damage detection by comparing body orientation between the robot and its model. The proposed method identifies damaged legs and updates the robot’s model for integration into its control system. Experiments on uneven terrain validate its robustness and computational efficiency.

%Multi-legged robots (MLRs) deployed on complex missions are prone to physical leg damage, which can alter their leg morphology from the original design, limiting task performance and mission success. However, with proper reconfiguration, these robots can continue to move by adapting their gait to accommodate new morphologies. Achieving this adaptability requires a robust algorithm that can detect damage and dynamically update the robot’s self-model. This paper introduces a robust self-modeling and damage identification algorithm that enables a robot to autonomously update its morphology in response to partial or complete leg loss. Our method detects and identifies damage in MLRs using only embodied recorded data from a low-cost IMU, enabling autonomous self-modeling without complex sensor arrays. Additionally, we introduce a novel FFT-based filter to handle time-inconsistent signals, allowing accurate damage identification by comparing the main body orientation data between the robot and its model. As a result, the algorithm accurately identifies damaged legs and generates a new model for integration into their guidance, navigation, and control system. Tested across various scenarios, including uneven terrain, this algorithm demonstrates computational efficiency and robustness for autonomous damage identification in MLRs.

\end{abstract}

\begin{IEEEkeywords}
Multi-legged Robot, Damage Identification, Self-modeling, Genetic Algorithm.
\end{IEEEkeywords}

\IEEEpeerreviewmaketitle

\section{Introduction}

\IEEEPARstart{T}{he} study of intelligent and adaptable multi-legged robots (MLRs) navigating unstructured environments is a rapidly advancing field \cite{arm2023scientific, gong2023legged, Spot, Anymal}. In these challenging environments, MLRs are prone to performance degradation and unexpected damage, which often cannot be immediately repaired. When this happens, the robot's Guidance, Navigation, and Control system, reliant on a model of the healthy robot, may become ineffective. Therefore, a comprehensive algorithm is essential, allowing the robot to autonomously detect and identify damage, thereby preventing mission failure. This algorithm should also provide the updated model of the damaged robot for use for recovery. In this work, we strictly define "damage" as instances of partially broken limbs or the complete loss of limbs or legs resulting in changes in the robot's morphology. We do not consider issues related to sensor malfunctions, or damage-induced hardware failures.

Researchers have explored fault detection in MLRs using three primary approaches: (i) in-system diagnostics, (ii) external diagnostics, and (iii) model-based diagnostics. In-system diagnostics rely on internal data, like motor IDs, allowing for fast operation without complex algorithms or costly sensors. However, this approach can only identify connection loss and is often unreliable for more nuanced damage scenarios. For example, Kim \textit{et al.} use motor ID pinging and a connector pin to gather diagnostic data \cite{kim2017snapbot}. External diagnostics, commonly applied in swarm robotic systems, rely on external sensors or environmental cameras to assess damage. However, achieving reliable results requires sufficient external sensors, which may be cost-prohibitive or impractical in field robotics. In \cite{kim2020snapbot}, a camera-based algorithm employing color detection and motor IDs identifies leg damage, while \cite{ozkan2021self} demonstrates how nearby robots can transport a disabled robot to a designated area.

Model-based diagnostics, by contrast, detect and identify damage by comparing the real system with a set of models to match the system's behavior. For instance, sensor-based methods use attitude sensors, touch switches, and estimation algorithms to derive leg dimensions in quadrupeds \cite{liang2010self}. Johnson \textit{et al.} leverage contact event sensors for leg identification and recovery, investigating broken leg detection, identification, and recovery in hexapods \cite{johnson2010disturbance}. Other examples include joint angle and tilt sensors for model-driven damage detection \cite{bongard2006resilient} and acoustic-based fault detection using Fast Fourier Transform (FFT) analysis \cite{chattunyakit2019self}. Particle Swarm Optimization (PSO) algorithms paired with inertial measurement units (IMUs) have shown promise in enabling self-recovery in damaged quadrupeds \cite{chattunyakit2019bio, chattunyakit2016pso}. However, PSO's reliance on numerous sensors can increase costs and complexity. Studies comparing PSO with genetic algorithms (GA) found PSO more effective for leg length identification, while GAs performed adequately for identifying missing links \cite{chattunyakit2016pso}.

GAs are extensively applied in damage identification for robotic and structural systems. For instance, Zhang et al. employed multi-objective genetic programming for automated failure detection \cite{zhang2006autonomous}. In structural systems, Tang et al. integrated GAs with probabilistic models and support vector machines for vibrational fault diagnosis in rolling bearings, though results were affected by noise interference \cite{tang2020fault}. To address this issue, Sahu \textit{et al.} introduced adaptive GAs with regression analysis to filter noise for structural damage detection \cite{sahu2020adaptive}. Modifying GA functions—selection, crossover, and mutation—has improved accuracy in identifying damage locations within railway bridges \cite{yang2022structural}, while other applications include optimizing joint angles and velocities in rehabilitation robots \cite{azar2021optimized}. GA has also been applied in MLR damage identification and recovery. For instance, \cite{kon2020gait} proposes a GA for generating new hexapod gaits after joint loss, though the algorithm sometimes produces unusable gaits due to an inefficient definition of the fitness function.

This paper presents a novel damage identification framework designed to detect damage in MLRs, accurately identify damaged legs within the system, and generate an updated model of the damaged robot. This new model can be seamlessly integrated into model-based controllers and damage recovery algorithms. To demonstrate the algorithm's efficacy, we tested it on a hexapod using a single, low-cost IMU sensor mounted on the main body. Our approach identifies damage by comparing roll, pitch, and yaw data between the damaged robot and a modular dynamic model of the system in the time domain. Given this comparison is affected by uncertainties such as noise and delays, the proposed algorithm develops an FFT-based preprocessing algorithm to minimize the effect of such phenomena, leading to precise self-modeling approach.

The proposed method proves to be both computationally efficient and accurate across multiple damage scenarios, including cases with missing legs and links. The fast and modular model employed in this study is adaptable and generates dynamics for a damaged robot through modifications to a single parameter, termed the morphology vector. The model has been validated in \cite{future:farghdani_model2}, demonstrating its superior speed. With the help of this fast simulation engine, we are able to identify the damaged legs in a six-legged robot with 24 degrees of freedom (DoFs) in less than 10 minutes. The major contributions of this work are as follows:

\begin{enumerate}

    \item Unlike previous approaches \cite{kon2020gait, manglik2016adaptive, pap2010optimization, park2013automated, bongard2006resilient}, our algorithm achieves accurate damage identification without the need for complex sensor arrays or external data.
    
    \item In contrast to methods such as \cite{bongard2006resilient}, our approach explicitly computes closed-form analytical dynamics equations for the damaged robot, as a by-product.

    \item To demonstrate its robustness, our algorithm has been successfully tested on data collected from the damaged robot walking in realistic and challenging environments, including slopes, sand, and rocky terrain.

\end{enumerate}

This paper is organized as follows: In Section \ref{chap5:method}, we present our robust self-modeling and damage identification algorithm. We investigate the accuracy of our algorithm in multiple case studies, including different damage scenarios in Section \ref{chap5:results}. Section \ref{chap5:conclusion} includes some concluding remarks.

\section{Problem Statement}

The morphology of a healthy MLR consists of a main body indexed by $b$, $N$ legs indexed by $i\in\{1, \,...\, ,N\}$, and the $i^{th}$ leg includes $n_i$ links indexed by $j\in\{1, \,...\, ,n_i\}$, so $l_{ij}$ represents Link $j$ of Leg $i$. The total number of robot's leg DoF is defined by $N_T:=\sum_{i=1}^{N}n_i$; hence, the system has $6+N_T$ DoF. Note that the number of links in the legs can vary, mainly due to capturing physical damage in the system. Here, we assume that the MLR is equipped with a low-cost IMU on the main body and servo motors at the joints. The joint servos are used to follow the gait, while the IMU is used to detect damage and identify the updated morphology.

When an MLR experiences physical damage, its morphology changes due to lost or malfunctioning limbs.  
To represent the change in the system after damage occurrence, we define the morphology vector $\EX$, as follows:
 \begin{align}\label{chap5:Ex}
    &\EX = \begin{bmatrix} \EX_{leg_1} & \cdots & \EX_{leg_i} & \cdots & \EX_{leg_N}      
    \end{bmatrix}\in\R^{N_T},  \\
    &\EX_{leg_i}= \begin{bmatrix} \ex_{i1} & \cdots & \ex_{ij} & \cdots & \ex_{in_i}      
    \end{bmatrix} \in\R^{n_i}, \\ 
    &\ex_{ij}:=\left\{ \begin{array}{rcl}
     1 & \mbox{if the link is present} \\ 
     0 & \mbox{if the link is absent  } 
    \end{array}\right.
 \end{align}
This is a binary vector, divided into $N$ segments corresponding to the robot’s legs. In the absence of damage, all elements of $\EX$ are one and if a link is missing, the corresponding element is set to zero. We name each binary element $\ex_{ij}$, the existence number of the link $l_{ij}$.   
%The morphology vector is a row vector of zeros and ones divided into $N$ segments corresponding to the legs of the healthy robot. This vector only includes ones, if the system does not experience any damage.

\begin{assumption}\label{chap5:as1}
    %To identify the new system morphology, we assume that we have access to a set of IMU measurements corresponding to the motion of the damaged robot while it is moving with the nominal gait designed for the healthy system.
    To identify the post-damage system morphology, we assume access to a set of IMU measurements recorded while the damaged robot walked with the nominal gait designed for the healthy system.
\end{assumption}

\begin{problem}\label{chap5:pr1}
    %Given Assumption \ref{chap5:as1}, find the new morphology $\EX_{dam}$ such that we minimize a measure of difference between the recorded and simulated IMU signals.
    Under Assumption \ref{chap5:as1}, determine the morphology vector of the damaged system $\EX^{dam}$ by minimizing a measure of difference between the recorded and simulated IMU signals.
\end{problem}

\begin{figure}[hbt!]
  \centering
  \includegraphics[width=.35\textwidth]{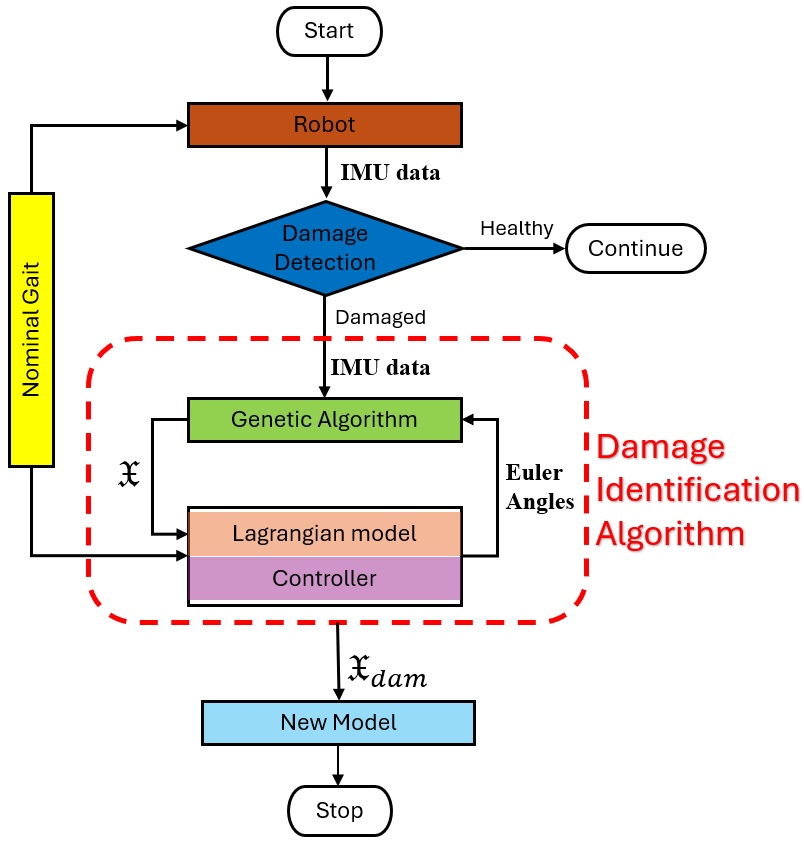}\\
  \caption{Schematic diagram of the proposed method}
  \label{fig:Schematic}
\end{figure}

A schematic diagram of the damage identification method proposed in this paper is shown in Fig. \ref{fig:Schematic}. The algorithm relies on an optimization engine based on a developed genetic algorithm. To numerically capture the damaged robot’s motion, we use the modular MLR simulation engine introduced in \cite{farghdani2024singularity, future:farghdani_model2}. This real-time engine utilizes a Boltzmann-Hamel Lagrangian formulation to model an MLR across various probable damage scenarios. The subsequent section provides a detailed explanation of the developed identification framework's components, enhancing the robustness and reliability of the technique. These include the GA-based optimization, signal processing techniques, and strategies for addressing the gap between simulation and reality caused by noise, unknown time delays, and inconsistent sampling rates.

%A schematic diagram of the proposed damage identification method is shown in Fig. \ref{fig:Schematic} that includes an optimization engine based on GA. To simulate the motion of a damaged robot in our algorithm, we use the modular MLR simulation engine from \cite{future:farghdani_model1, future:farghdani_model2}. This real-time engine works with a Boltzmann-Hamel Lagrangian model to simulate the motion of an MLR with physical leg damage.
%The following sections provide a detailed explanation of each component. The rest of this section details the adoption of GA for damage identification of MLRs and the conditioning required to enable comparison of simulated and experimented motion addressing noise, unknown time delay, and inconsistent sampling rate (sim-to-real gap).

\begin{remark} 
    The proposed algorithm relies solely on proprioceptive sensing of the robot’s motion, with no use of external sensors, making it a fully embodied self-identification method. 
\end{remark}

\begin{remark} 
    We use a \textit{single} set of IMU measurements recorded while the damaged MLR is walking with the nominal gait. We avoid random system excitation (a common practice in system identification techniques) in both simulations and experiments to prevent further damage to the robot. 
\end{remark}

%\begin{remark} Our algorithm assumes only proprioceptive sensing of the robot's motion and no external sensors; hence, considered as an embodied self-identification method. \end{remark}

%\begin{remark} We only use \textit{one} set of IMU measurements for the damage system moving with nominal gait and we avoid random excitation of the system (common practice in system identification techniques) both in simulations and experiments that can further damage the system. \end{remark}

\section{Methodology}\label{chap5:method}

To address Problem \ref{chap5:pr1}, this section details our proposed damage identification algorithm for MLR systems. 

\subsection{Genetic Algorithm for MLRs}

In this paper, we develop a binary search genetic algorithm with population size $P_s$, and generation size $G_s$ to obtain the morphology of a damaged MLR. The input to the algorithm is a set of recorded robot's main body orientations while walking with the nominal gait, and the output is the morphology vector corresponding to the lowest value of the cost function, whose form is introduced later. Candidate $k$ in the population is specified by the morphology vector $\EX^k$, whose elements distinguish between a healthy and damaged leg link in the robot. To ensure that every candidate generated by the GA optimization corresponds to a physically feasible morphology, we must introduce a refining algorithm, named link logic. Given a candidate $\EX^k$, the algorithm starts by assessing the existence number of the links immediately connected to the main body, i.e., $\ex_{i1}$ for $i\in\{1,\cdots,N\}$. For Leg $i$, if $\ex_{i1}$ is found to be $0$, the algorithm shall set $\ex_{ij}=0$ for all following links, i.e., for all $1<j\leq n_i$, since these links cannot exist in the absence of the first leg link. Otherwise, when the existence number $\ex_{i1}$ is $1$, the algorithm will repeat the same logic for $\ex_{i2}$, and so on. This process will continue until the last link of each leg. Below is an example of how the link logic algorithm works for a 3 DoF leg.
\begin{equation}\nonumber
 \mbox{GA generated}: \EX_{leg_i}=[1 ~ \textbf{0 1}] \Rightarrow \mbox{After Link Logic}: [1 ~ \textbf{0 0}]
\end{equation}
We summarize the link logic algorithm in Algorithm \ref{Link Logic}.
\begin{algorithm}
\caption{Link logic for an MLR}
\begin{algorithmic}[1]\label{Link Logic}
\REQUIRE Number of legs $N$ and the DoF of each leg $n_i$
\REQUIRE Candidate morphology vector $\EX^k$ in the form of Eq. \eqref{chap5:Ex}
\FOR{$i = 1$ to $N$}
    \FOR{$j = 1$ to $n_i$}
        \IF{$\ex_{ij} = 0$}
            \FOR{$\kappa = j+1$ to $n_i$}
                \STATE $\ex_{i\kappa} = 0$
            \ENDFOR
        \ENDIF
    \ENDFOR 
\ENDFOR
\RETURN Refined candidate morphology vector $\EX^k$
\end{algorithmic}
\end{algorithm}

\subsubsection{Initialization}

The initial population is created by randomly generating morphology vectors of size $\mathbb{R}^{N_T}$. For each candidate, the algorithm applies the link logic in Algorithm \ref{Link Logic} to ensure a physically feasible connected system. 
%Specifically, if the first link of a leg existence number is assigned $0$, all subsequent elements ($\ex_{ij}$, $1<j\leq n_i$) corresponding to that leg are set to $0$, as no further links can exist without the preceding ones. Conversely, if the first link existence number is assigned $1$, the process continues iteratively for the remaining links until the last link of the leg.
The algorithm ensures that all candidates are unique to maintain diversity within the population. If two candidates are found to be identical after the link logic correction, the randomization process is repeated until we have $P_s$ number of unique candidates for the initial population. 
%Once the candidates are finalized, they are passed to the simulation engine to generate the trajectory of the robot's main body orientation data, which serves as input for the subsequent steps of the algorithm.

\begin{comment}
    
\subsubsection{Crossover}

The crossover is a sequential procedure starting from the best fit morphology. Once the candidates are arranged in terms of an ascending cost function, the second half of the candidates with higher cost function undergo crossover with the candidates with lower cost function. This is shown in Eq. \eqref{crossover}, where each link in a candidate is crossed over with that of a random candidate with a lower cost function to create a greater difference among the candidates. 
\begin{equation}\label{crossover}
    x_{k,ij} = x_{random(1:k-1),ij}
\end{equation}
The link logic in this section is embedded within the generation of the population same as the initialization step. This is done by first applying Eq. \eqref{crossover} to the first link, and, if its existence number is $0$, then setting the next links to be $0$ as well. If the existence number is $1$, then Eq. \eqref{crossover} is applied to the next link and continues until the last link. This helps selectively choose the crossover link morphology and preserves the crossover information, which would have been changed if the link logic had been applied afterward.
\end{comment}

\subsubsection{Evaluation}

Once a population is generated, each candidate's morphology is fed into the simulation engine to produce the trajectory of the main body orientation data over a period of time. The experimental and simulated data are then passed through a novel filter (see Section \ref{fft_filter}) to evaluate the walking performance of the robot compared to real-world data collected prior to the damage identification phase. This filter addresses several bottlenecks affecting the comparison of the simulated and experimental data, including noise, drift, time delay, and inconsistency in sampling frequency.

The cost function $\mathcal{F}_k$, determined for the morphology candidate $\EX^k$, is defined to be the accumulated summation of the absolute error in the main body orientation over time. We claim that a small value of $\mathcal{F}_k$ corresponds to a candidate MLR that behaves similarly to the damaged robot.
To parameterize the orientation, we extract Euler angles—roll, pitch, and yaw—from the rotation matrix data. We aim to identify the optimal candidate morphology $\EX^{dam}$ that minimizes the discrepancy between the simulated and experimental orientation data. The problem is formulated as: 
\begin{align}
     \EX^{dam}&=\arg\!\min_{\EX^k \in \mathcal{K}} \mathcal{F}_k, \label{GA:fitness} \\
    \text{such that} ~ \mathcal{F}_k&= \sum_{q=1}^\mathcal{Q} \Big( 
    \vert \Phi^k_{\text{Sim}}(t_q) - \Phi_{\text{Exp}}(t_q) \vert + \notag \\
     & \vert \Theta^k_{\text{Sim}}(t) - \Theta_{\text{Exp}}(t_q) \vert + 
    \vert \Psi^k_{\text{Sim}}(t_q) - \Psi_{\text{Exp}}(t_q) \vert \Big), \notag
\end{align}
where:
\begin{itemize}
    \item \( \mathcal{F}_k \) is the cost function for candidate \( k \), representing the total orientation error over time.
    \item \( \Phi^k_{\text{Sim}}(t_q) \), \( \Theta^k_{\text{Sim}}(t_q) \), and \( \Psi^k_{\text{Sim}}(t_q) \) are respectively the simulated roll, pitch, and yaw angles at the time sample $t_q$ for the candidate morphology $\EX^k$.
    \item \( \Phi_{\text{Exp}}(t_q) \), \( \Theta_{\text{Exp}}(t_q) \), and \( \Psi_{\text{Exp}}(t_q) \) are respectively the roll, pitch, and yaw angles obtained experimentally at  \( t_q \).
    \item \( \mathcal{Q} \) is the number of time samples.
    \item \( \mathcal{K} \) represents the set of all feasible morphologies.
\end{itemize}

\subsubsection{Crossover}

For each GA generation, the crossover process works based on the identification of the best-fit morphologies. In a population, we rank all candidates in descending order according to their value of the cost function. We keep the first half—those with lower cost values—unchanged and change the morphology candidates in the second half—those with higher cost values—according to the following rule:
\begin{equation} \label{ga_crossover}
       \ex^k_{ij} = \begin{cases} 
            \ex^{\text{rand}(1:P_s/2)}_{ij}, & \text{if } r_c^k \leq CR \\
            \ex^k_{ij} , & \text{otherwise}
        \end{cases}
\end{equation}
where $r_c^k$ is a randomly generated number between zero and one, $CR \in [0,1]$ is the crossover rate, and $\text{rand}(1:P_s/2)$ indicates a random candidate in the first half of the population in the crossover process. According to Eq.~\eqref{ga_crossover}, there is a chance for each link in the second half candidates to be replaced with the corresponding link from a randomly selected candidate with a lower cost value, to enhance diversity within the population. The link logic in Algorithm \ref{Link Logic} is then integrated directly into the newly generated candidates during the crossover process.

\subsubsection{Mutation}

The mutation process is performed after crossover and is applied independently to each leg of the newly generated candidates. Given the morphology vector $\EX^k_{leg_i}$ for Leg $i$, we mutate the existence numbers of the leg links only if we have $r_m^k \leq MR$, where $r_m^k$ is a randomly generated number between zero and one, and $MR \in [0,1]$ is the mutation rate.  
Mutation for a selected leg with index $i$ means that each link existence number $\ex^k_{ij}$ is modified to $1$ or $0$ based on predefined probabilities:
\begin{equation} \label{ga_mutation}
       \ex^k_{ij} = \begin{cases} 
            1, & \text{if } \bar{r}_m^k \leq P_{ij} \\
            0 , & \text{otherwise}
        \end{cases}
\end{equation}
The probability of assigning a value of $1$ to Link $j$ of Leg $i$ is denoted as $P_{ij} \in [0,1]$ and $\bar{r}_m^k\in [0,1]$ is a random number. These probabilities are designed to be higher for links closer to the main body, reflecting the assumption that damage is less likely in proximal links compared to distal ones. This approach also increases the likelihood of including the distal links in the morphology, as their presence depends on the existence of preceding links. The probabilities $P_{ij}$ are determined through a combination of estimation and experimental validation to balance algorithm accuracy with computational efficiency. By favoring morphologies that align with real-world damage patterns, the mutation process helps refine the candidate pool and improves the likelihood of convergence within a reasonable number of iterations. Similar to the crossover step, the link logic is enforced during mutation to ensure physically valid morphologies. 

\subsubsection{Selection}

In this paper, a simple selection method is applied: keep the first half of the population unchanged and replace the second half with new candidates generated during crossover and mutation.  
\subsubsection{Termination} 

The GA optimization process repeats until a stopping criterion is met. This could be the maximum number of iterations or a convergence criterion.

\subsection{Fourier Transform Filter}\label{fft_filter}

To perform the optimization described in the previous section, it is necessary to compare experimental data obtained from a low-cost IMU mounted on the main body of the MLR with corresponding signals generated by the simulation. However, direct signal comparison proves ineffective due to several factors:

\begin{enumerate} 
\item Sensor noise and drift inherent to the IMU, which directly influence the cost function, 
\item Inconsistent time delays within the experimental setup, leading to random temporal offsets in the recorded data, 
\item Variability in the IMU's sampling rate, fluctuating between 950 Hz and 1 kHz, which introduces inaccuracies in time labeling, and
\item Uncontrolled environmental factors, such as uneven or slippery terrain, that affect the system dynamics. \end{enumerate}
% The algorithm from the previous section was developed and tested on a six-legged robot; however, its performance was not satisfactory due to several issues. 
% First, the experimental data was noisy, directly affecting the cost function value. Second, there was a time delay between the motion control commands sent to the robot and the actual start of its movement. This delay was inconsistent and varied across tests, causing a time shift in the recorded data from the robot. Third, due to the poor quality of our IMU sensor, the sampling rate was not constant during the tests, fluctuating between $950\,Hz$ and $1000\,Hz$. This inconsistency impacted the time labels of the data, as a uniform sampling rate was assumed for the entire signal. Finally, sensor drift and the uneven ground of the experimental setup resulted in different initial conditions between the simulation and the experiment. 
These factors can occasionally result in scenarios where the correct leg morphology produces a higher cost function value than an incorrect one, particularly under certain damage conditions.

To address the aforementioned issues, we propose a Fast Fourier Transform (FFT)-based filter to preprocess both experimental and simulated IMU signals before their comparison in the cost function \eqref{GA:fitness}. In the first step, both orientation signals are transformed into the frequency domain to extract their Power Spectra (PS) corresponding to the existing frequencies. Note that for an arbitrary signal $\mathbb{X}$ of length $n$ in time domain, its FFT $\mathbb{Y}$ is defined by \cite{frigo1998fftw}:
\begin{align}  \label{fft}
    \mathbb{Y}(\alpha) = \sum_{\beta=1}^{n} \mathbb{X}(\beta) W_n^{(\beta-1)(\alpha-1)}, 
\end{align}
accordingly the Inverse FFT (IFFT) equation is as follows:
\begin{align} 
    \label{ifft}
    \mathbb{X}(\beta) = \frac{1}{n} \sum_{\alpha=1}^{n} \mathbb{Y}(\alpha) W_n^{-(\beta-1)(\alpha-1)}.
\end{align}
Here, $W_n = e^{\frac{-2\pi i}{n}}$ is one of the $n$ roots of unity.
Next, we define a cutoff frequency ($F_c$) and a PS threshold ($P_c$) for our filter. We filter higher than $F_c$ frequencies by setting the power associated to those frequencies equal to zero in the PS of the orientation signals, since the MLR's gaits often move the robot at low frequencies.  
We further retain only the peaks with power values above $P_c$ in roll and pitch signals, which typically exhibit oscillatory behavior, and the rest of the nonzero PS are set to zero. This process emphasizes that continuous roll and pitch motion in one direction, which could lead to a robot ``flip", is not expected. For the yaw angle, however, non-oscillatory motion may occur in certain damage scenarios. Therefore, for the yaw angle, all frequencies with power values greater than $P_c$ are preserved. The final step involves signal reconstruction. Before applying IFFT on the conditioned signals, we define a consistent time vector and common initial conditions for both simulated and experimental signals. This process mitigates any mismatches in sampling rates and ensures motion synchronization. An overview of the FFT-based filter algorithm is illustrated in Algorithm 2. 
\begin{algorithm} 
\caption{FFT-based Filter}
\begin{algorithmic}[1]
\REQUIRE Trajectory of the main body orientation data

\STATE Extract PS of data based on \eqref{fft}
\STATE Remove frequencies higher than $F_c$
\FOR{ Roll $\&$ Pitch angles}
    \STATE Identify frequencies corresponding to power peaks greater than $P_c$
    \STATE Set the power corresponding to rest of the frequencies equal to zero
\ENDFOR
\FOR{ Yaw angle}
    \STATE Identify all frequencies with power greater than $P_c$ 
    \STATE Set the power corresponding to rest of the frequencies equal to zero
\ENDFOR
\STATE Reconstruct Euler angles trajectory based on \eqref{ifft}
\RETURN Reconstructed trajectory of the main body orientation data
\end{algorithmic}
\end{algorithm}

\begin{figure}[hbt!]
  \centering
  \includegraphics[width=.35\textwidth]{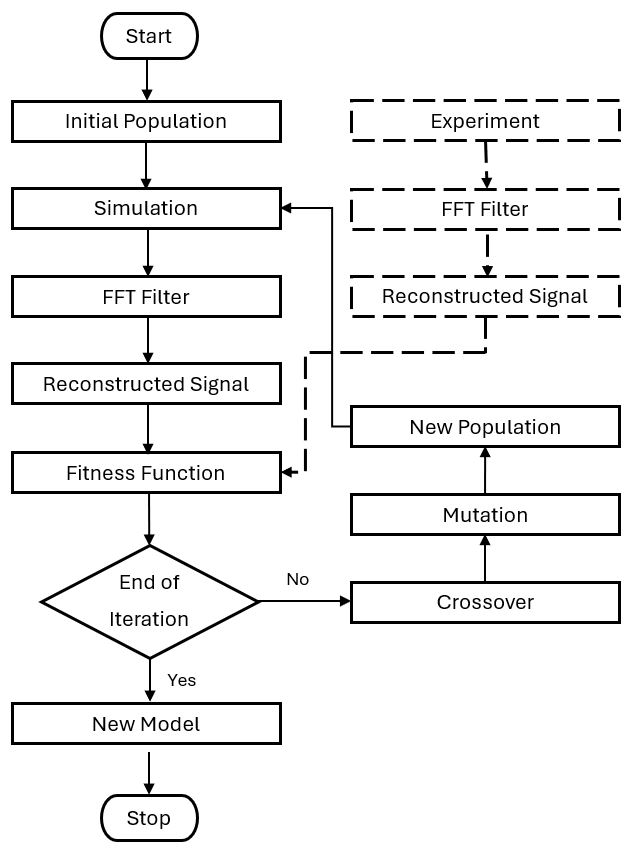}\\
  \caption{Damage identification algorithm flowchart}
  \label{fig:ga}
\end{figure}

At the outset of this section, we identified three primary challenges—sensor noise, time delay, and variable sampling rate—which are effectively addressed by the proposed filtering framework. First, as the robot's motion primarily resides in the low-frequency spectrum, high-frequency noise can be effectively attenuated by selecting an appropriate cutoff frequency, $F_c$. This filtering step isolates meaningful motion-related components while suppressing unwanted disturbances. Second, during signal reconstruction, both experimental and simulated signals are temporally aligned to share a common starting point, thereby eliminating relative time delays. Furthermore, the damage detection algorithm, introduced in the next section, enhances temporal synchronization through additional corrective measures.
Third, a unified time vector is defined during signal reconstruction to ensure that both experimental and simulated signals share a consistent sampling rate. This harmonization facilitates accurate and reliable signal comparison.
An overview of the damage identification algorithm, integrating the FFT-based filter, is presented in Fig.~\ref{fig:ga}.

\section{Results \& Discussion}\label{chap5:results}

We evaluate the performance of the proposed damage identification method for MLRs using the algorithm introduced in Section~\ref{chap5:method}. As a representative case study, we consider a six-legged robot modeled as a multibody system, where each leg possesses 3 DoF. The effectiveness of the method is demonstrated through extensive experimentation across various damage scenarios, encompassing both ideal and challenging environmental conditions, as detailed in the subsequent sections.

\subsection{Platform Description}

\begin{figure}[hbt!]
  \centering
  \includegraphics[width=.3\textwidth]{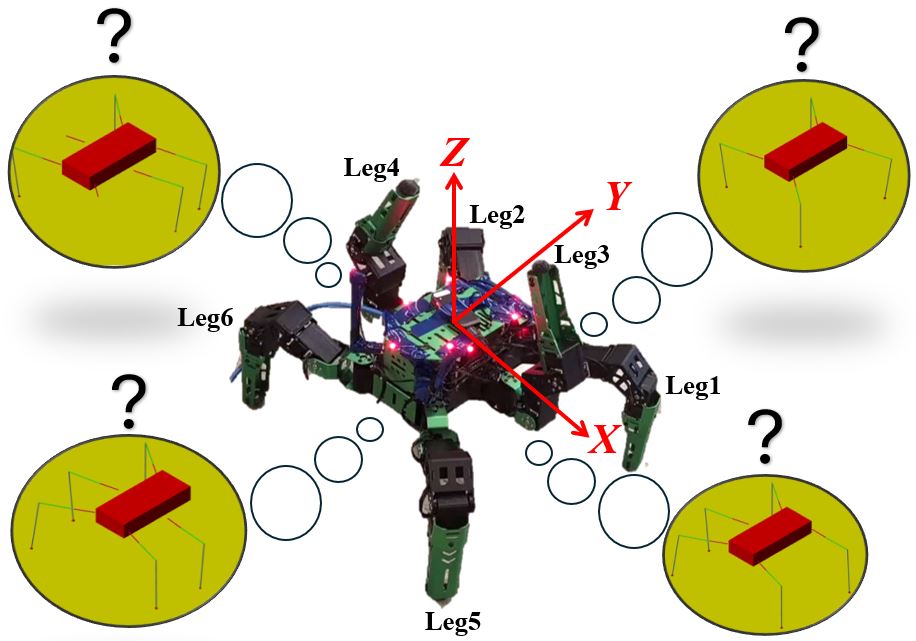}\\
  \caption{Outline of the algorithm and experimental setup}
  \label{Robot}
\end{figure}

The experimental setup incorporates a Hiwonder JetHexa, a six-legged robot with 3-Dof legs. Each leg has three serial bus servo motor-actuated joints. The onboard robot computer is an NVIDIA Jetson Nano Developer Kit with an MPU6050 IMU \cite{jethexa}. Fig. \ref{Robot} shows a picture of the system. This IMU is used to obtain the roll, pitch, and yaw of the robot's body, and it is the only sensor data used in this paper. 
In all our experimental tests and simulations, we utilize the model, controller, and trajectories introduced in \cite{ farghdani2024singularity, future:farghdani_model2}. The gait trajectory is specifically designed to enable a healthy robot to walk in the $Y$ direction using a tripod gait. In this gait, the legs are divided into two synchronized groups: first group consists of Legs $\lbrace 1, 4, 5 \rbrace$, while the second includes Legs $\lbrace 2, 3, 6 \rbrace$.

\subsection{Damage Detection}\label{detection}

In the literature, various sensors mounted on the robot, such as touch sensors \cite{1308802}, encoders\cite{5651061}, and cameras\cite{6618086}, have been employed for damage detection in MLRs \cite{buettner2022review}. However, in our approach, we focus exclusively on the data from a single IMU sensor located on the robot's main body, which inherently limits the available options for analysis. To detect damage, we employ a sliding window detector, a method previously applied to sensor fault detection in wheel-legged robots \cite{10489305}. This technique leverages a temporal window to analyze fluctuations in sensor signals, offering an effective way for identifying anomalies within a constrained dataset.

Under nominal conditions, a healthy robot is typically designed to minimize roll and pitch motions during locomotion. Therefore, deviations in these angles can serve as indicators of potential damage. To quantify these deviations, we implemented a sliding window method with a 0.5-second window size, approximately corresponding to half of a gait cycle. Within each window, the maximum and minimum values of the roll and pitch angles were identified, and their difference was used to measure the fluctuation in these signals.
For a healthy robot, this fluctuation remains below 5 degrees. If the fluctuation exceeds this experimentally determined threshold and continues for 2 seconds, the system flags the presence of damage. To improve the resolution and accuracy of detection, the moving window is updated every 0.1 seconds, providing finer temporal granularity in identifying anomalies. This detector can also assist in identifying the motion's starting moment, thereby minimizing the impact of initial delays. 

\begin{figure}[hbt!]
  \centering
  \includegraphics[width=.4\textwidth]{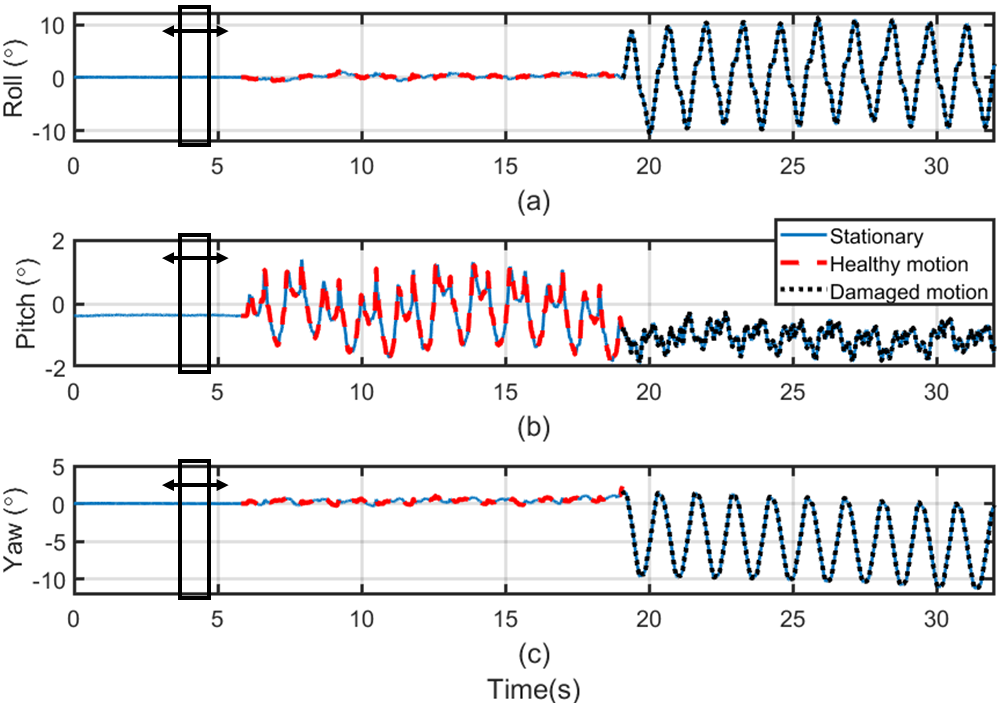}\\
  \caption{Damage detection algorithm output from the robot main body orientation data}
  \label{fig:detection}
\end{figure}

To demonstrate the effectiveness of our damage detection method, we analyzed the roll, pitch, and yaw angle of the robot's main body over 32 seconds. During the first 6 seconds, the robot remains stationary, with the recorded signal primarily reflecting noise. From 6 to 19 seconds, the robot operates in a healthy state. At $t=19$, damage occurs, specifically affecting Legs 3 and 4, resulting in losing them completely. Although the robot maintains stability using its remaining four legs, its motion deviates significantly from the intended design. Since the pitch angle remains below 2 degrees in this damage scenario, it does not serve as a reliable reference for damage detection. As shown in Fig. \ref{fig:detection}, the damage detector successfully identifies both the occurrence of the damage and the motion's starting moment with acceptable accuracy.

\subsection{Damage Identification}\label{identification}

In this section, the results of the Fig. \ref{fig:ga} algorithm are demonstrated and discussed in detail. Table \ref{table:ga} summarizes the parameters used for the GA algorithm and FFT-based filter.
\begin{table}[ht]
\caption{Damage identification algorithm parameters} 
\color{black}
\centering 
\begin{tabular}{c c c c c } 
\hline\hline 
  Parameter & Value &  Parameter & Value\\ 
\hline 
$P_s$ & 10 & $G_s$ & 20 \\
CR & 0.9 & MR & 0.33 \\
$P_{i1}$  & 0.9 &  Simulation time & 5 sec \\
$P_{i2}$ & 0.7 & $P_{i3}$ & 0.7 \\
$F_c$ & 10 Hz & $P_c$  & 0.1 \\
\\
\hline 
\end{tabular}
\label{table:ga} 
\end{table}

The results of the damage identification algorithm for 18 tests across 8 different damage scenarios are summarized in Tables \ref{tab:results} and \ref{tab:results2}. Table \ref{tab:results} compares the actual and predicted morphologies, highlighting the algorithm's performance. The first column represents the actual damage scenarios, while the rest of the rows and columns represent the predicted scenarios based on the morphology vectors of each leg. 

To provide a clearer interpretation of the results, the damage scenarios can be categorized into two groups: single-leg damage and double-leg damage. Table \ref{tab:results2} shows that the algorithm achieved an average convergence time of 10 minutes and an overall accuracy of $89\%$, with higher success rates in double-leg damage scenarios compared to single-leg damage. In double-leg damage scenarios, the algorithm reliably identified the damaged legs but struggled to determine the number of missing links. This limitation stems from several factors. First, during experiments, the robot’s links were not fully detached but instead locked at angles that did not affect the general limping motion, whereas the simulation completely removed the links. This discrepancy introduced an unmodeled additional mass between the simulation and experimental setups. Second, the robot’s geometry and gait trajectory caused legs with missing links to often fail to touch the ground, particularly when the last two links were missing, making it challenging for the algorithm to detect their presence using only the IMU data from the main body. Third, the small mass of the links relative to the main body rendered their effects on the IMU measurements negligible unless they made ground contact. As a result, the algorithm could not accurately determine the number of missing links in damaged legs.

For single-leg damage scenarios, the algorithm demonstrated some inaccuracies at the leg level. In a limited number of tests, damage was predicted in a healthy leg from the same leg group. These results suggest that the algorithm is conservative, often overestimating damage within the same group. This behavior can be attributed to the robot’s tripod gait, where Legs $\lbrace 1,4,5 \rbrace$ and Legs $\lbrace 2,3,6 \rbrace$ alternate between the swing and support phases. Damage to one leg in a group compromises the group’s stability during the support phase, which, in turn, affects the swing phase of the opposing group.

%Comparing motion patterns between scenarios with one damaged leg in each group and two damaged legs in one group reveals significant differences. In scenarios where one leg is damaged in each group, the robot oscillates around its initial position in the X and Z directions while maintaining its position in the Y direction. The damaged legs remain in contact with the ground, preventing forward motion and limiting the observed effects to the swing phase. Conversely, in scenarios where two legs are damaged in one group, the robot exhibits forward motion in the Y direction and lateral motion in the X direction. For instance, if Legs 2 and 3 are damaged, the second group $\lbrace 2,3,6 \rbrace$ cannot maintain stability during its support phase, resulting in a drop in the Z direction. This disrupts the swing phase of the first group $\lbrace 1,4,5 \rbrace$. Additionally, the remaining leg in the second group (Leg 6) causes rotational motion about the Z axis during the support phase, further contributing to motion in the X and Y directions.

Comparing motion patterns between scenarios with one damaged leg per group and two damaged legs in one group reveals notable differences. When one leg is damaged in each group, the robot oscillates around its initial X and Z positions while remaining stationary in Y. The damaged legs stay grounded, blocking forward motion and limiting effects to the swing phase. In contrast, when two legs are damaged in one group, the robot moves forward in Y and laterally in X. For example, if Legs 2 and 3 are damaged, the second group $\lbrace 2,3,6 \rbrace$ loses support-phase stability, causing a Z-direction drop and disrupting the first group’s $\lbrace 1,4,5 \rbrace$ swing phase. The remaining leg (Leg 6) also induces Z-axis rotation, adding to the X and Y motion.

%In single-leg damage scenarios, the robot’s motion resembles that of two damaged legs in one group but with reduced severity. For example, if only Leg 3 is damaged, the second group $\lbrace 2,3,6 \rbrace$ has more legs to stabilize the robot, yet stability is still compromised during the support phase. This results in a smaller drop in the Z direction compared to double-leg damage, while translational motion in the X and Y directions is greater. Furthermore, two legs remaining in the second group, create a bigger rotational motion about the Z axis during the support phase. As we only focus on the orientation of the main body for damage identification, translational motion does not enter the algorithm. In addition, due to slippage, the yaw motion observed in single-leg damage scenarios is less than expected in the experimental tests, sometimes making it similar to simulated double-leg damage scenarios and complicating the algorithm's ability to differentiate between these cases.

In single-leg damage scenarios, the robot’s motion resembles that of double-leg damage in one group but with reduced severity. For instance, if only Leg 3 is damaged, the second group $\lbrace 2,3,6 \rbrace$ retains more legs for stabilization, though support phase stability is still impaired. This leads to a smaller Z-direction drop than in double-leg damage, but greater X and Y translational motion. The two remaining legs also induce larger rotational motion about the Z axis during support. Since the algorithm focuses solely on body orientation, translational motion is excluded. Additionally, due to slippage, yaw motion in single-leg damage is often less pronounced in experiments, making it resemble simulated double-leg cases and complicating differentiation.
 
In conclusion, while the algorithm's inaccuracy in single-leg damage scenarios highlights its conservative nature, it consistently provides reliable information about the occurrence of damage within the leg group. This ensures that no damaged leg is missed, even at the cost of occasional misidentification within the same leg group.
\begin{table}[h!]
\centering
\caption{Damage identification algorithm results}
\label{tab:results}
\resizebox{\columnwidth}{!}{%
\begin{tabular}{|l|c|c|c|c|c|c|}
\hline
\textbf{Actual/Predicted} & $\mathbf{\EX_{Leg1}}$ & $\mathbf{\EX_{Leg2}}$ & $\mathbf{\EX_{Leg3}}$ & $\mathbf{\EX_{Leg4}}$ & $\mathbf{\EX_{Leg5}}$ & $\mathbf{\EX_{Leg6}}$  \\ \hline
\textbf{Leg 3 missed}  & [1 1 1]  & [1 1 0]  & [0 0 0] & [1 1 1]  & [1 1 1]  & [1 1 1]  \\ \hline
\textbf{Leg 5  missed}  & [1 1 1]  & [1 1 1]  & [1 1 1] & [1 1 1]  & [1 0 0]   & [1 1 1]  \\ \hline
\textbf{Legs 1 \& 4 missed}  & [1 1 0]  & [1 1 1]  & [1 1 1] & [1 0 0]  & [1 1 1]   & [1 1 1]  \\ \hline
\textbf{Leg 3 \& 4 missed}  & [1 1 1]  & [1 1 1]  & [1 0 0] & [1 1 0]  & [1 1 1]   & [1 1 1]  \\ \hline
\textbf{Leg 3 \& 4 two last links missed}  & [1 1 1]  & [1 1 1]  & [1 1 0] & [1 1 0]  & [1 1 1]   & [1 1 1]  \\ \hline
\textbf{Leg 3 \& 4 one last links missed}  & [1 1 1]  & [1 1 1]  & [1 1 0] & [1 1 0] & [1 1 1]   & [1 1 1]  \\ \hline
\textbf{Legs 4 \& 5 missed}  & [1 1 1]  & [1 1 1]  & [1 1 1] & [1 0 0]  & [1 0 0] & [1 1 1] \\ \hline
\textbf{Legs 2 \& 3 missed}  & [1 1 1]  & [0 0 0]  & [1 1 0] & [1 1 1]  & [1 1 1]   & [1 1 1]  \\ \hline
\end{tabular}%
}
\end{table}

\begin{figure}[hbt!]
  \centering
  \includegraphics[width=.35\textwidth]{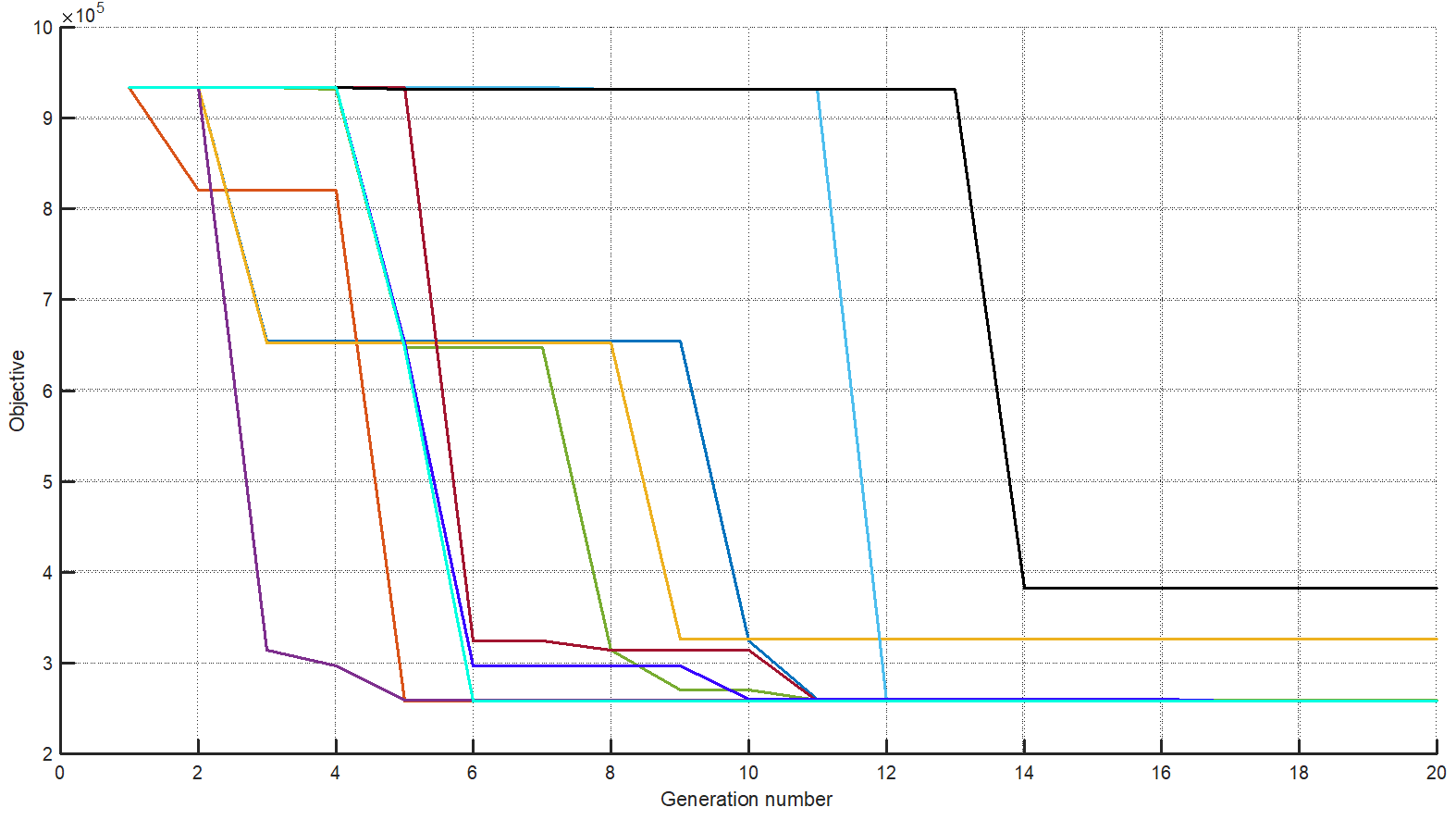}\\
  \caption{The best objective function value per generation for the scenario where Legs 4 and 5 are missing}
  \label{fig:j_plot}
\end{figure}

\begin{table}[h!]
\centering
\caption{Summary of Damage Identification Results}
\label{tab:results2}
\resizebox{\columnwidth}{!}{
\begin{tabular}{@{}lccc@{}}
\toprule
\textbf{Scenario} & \textbf{Tests Conducted} & \textbf{Correct} & \textbf{Time (min)} \\ 
%\textbf{Scenario} & \textbf{Tests Conducted} & \textbf{Correct} & \textbf{Accuracy (\%)} \\ 
\midrule
Leg 3 missed    & 2  & 1  & 9.47 \\ 
Leg 5  missed   & 2  & 1  & 9.62  \\ 
Legs 1 \& 4 missed    & 4  & 4  & 9.41 \\ 
Leg 3 \& 4 missed    & 2  & 2  & 9.73 \\ 
Leg 3 \& 4 two last links missed & 2  & 2  & 9.38 \\ 
Leg 3 \& 4 one last links missed & 2  & 2  & 9.81  \\ 
Legs 4 \& 5 missed & 2  & 2  & 9.06 \\ 
Legs 2 \& 3 missed & 2  & 2  & 9.44  \\ 
\midrule
\textbf{Overall} & \textbf{18} & \textbf{16} & \textbf{9.48} \\ 
\bottomrule
\end{tabular}
}
\end{table}
To further validate the algorithm's convergence and robustness, we repeated the damage identification process 10 times for each test scenario. As an example, the best objective function value per generation for the Legs 4 and 5 missing scenario is shown in Fig. \ref{fig:j_plot}. Across the 10 identification runs, the resulting morphology was either identical or differed by a single link within the identified damaged legs, a discrepancy discussed before. The most frequently identified morphology is reported in Table \ref{tab:results2}, representing the most probable morphological configuration based on the algorithm's convergence behavior. As shown in Fig. \ref{fig:j_plot}, the algorithm converged to three distinct morphologies over 10 attempts. In 8 runs, the identified topology indicated that Legs 4 and 5 were completely missing. Once the algorithm converged to a morphology where Leg 4 was entirely missing, and the last two links of Leg 5 were absent. In another attempt, Leg 5 was completely missing, while the last two links of Leg 4 were absent.

\subsection{Harsh Environment}\label{harsh}

\begin{figure}[hbt!]
  \centering
  \includegraphics[width=.35\textwidth]{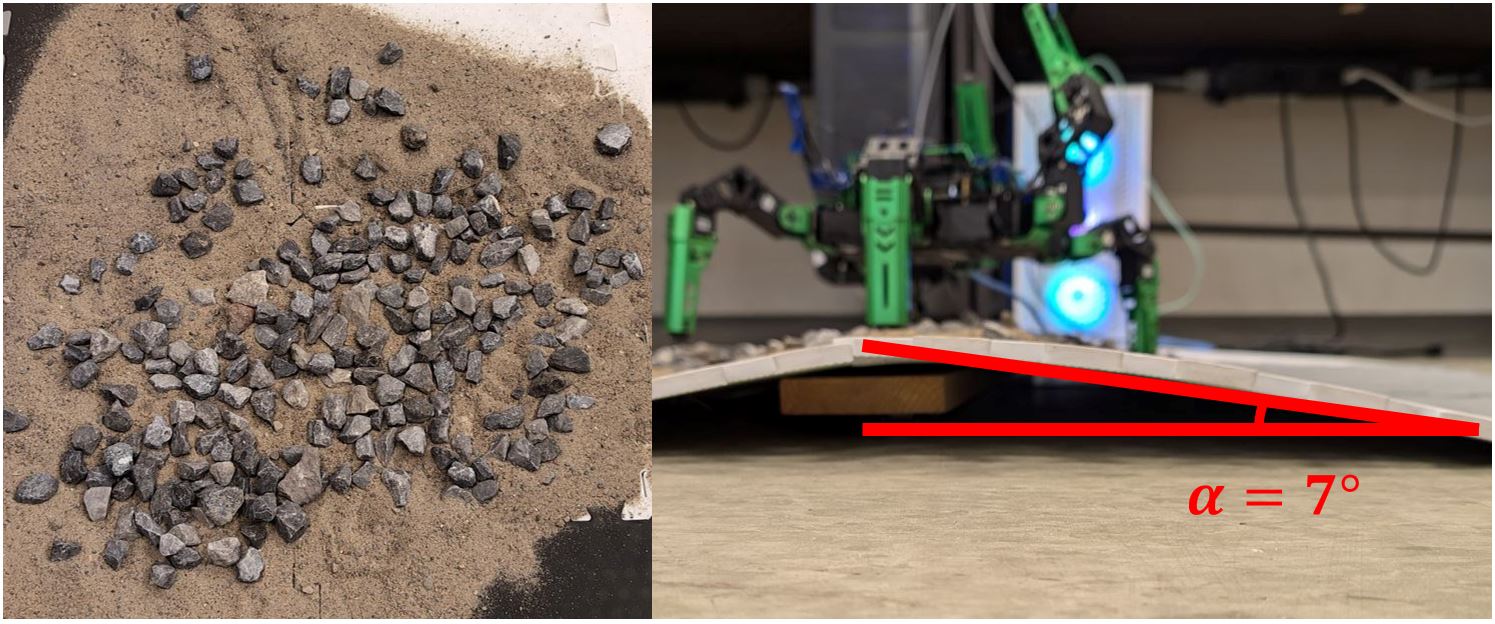}\\
  \caption{Harsh environment test setup }
  \label{fig:harsh}
\end{figure}

\begin{figure}[hbt!]
  \centering
  \includegraphics[width=.4\textwidth]{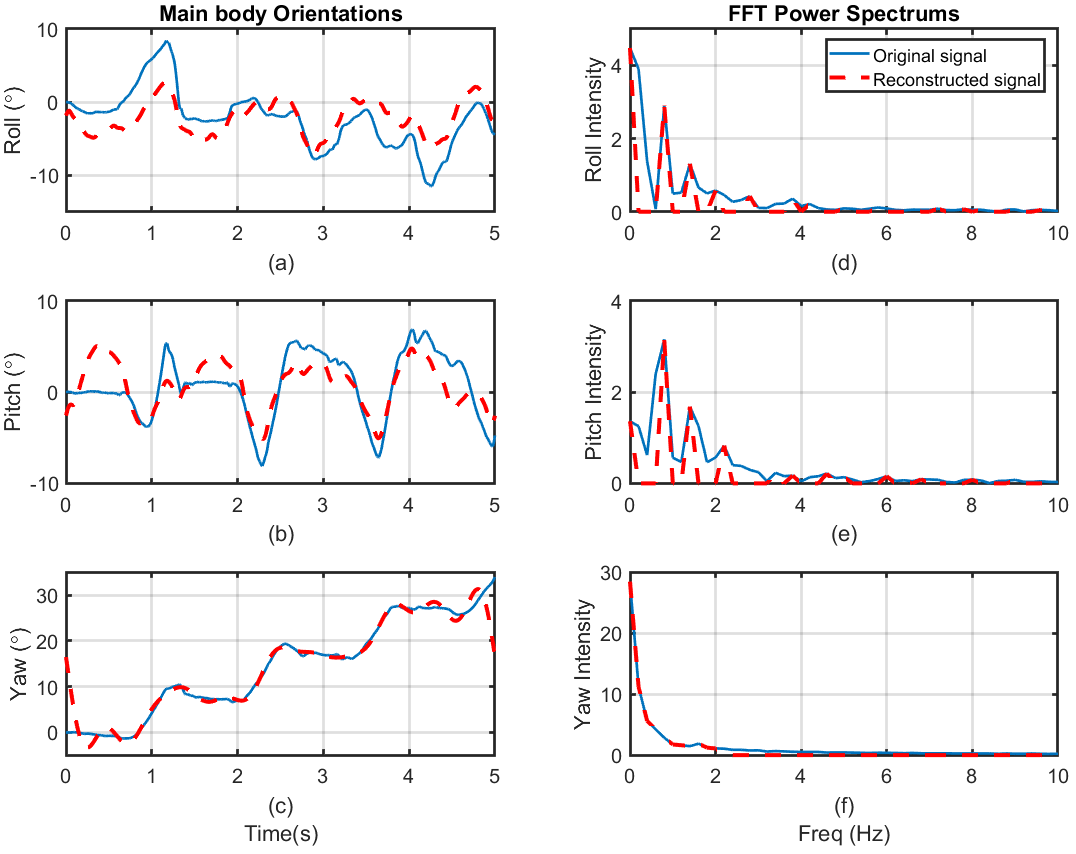}\\
  \caption{Robot main body orientation experimental data: (a), (b), (c) Original and filtered signals in time domain; (d), (e), (f) Original and filtered PS plots in frequency domain}
  \label{fig:harsh2}
\end{figure}

\begin{figure}[hbt!]
  \centering
  \includegraphics[width=.45\textwidth]{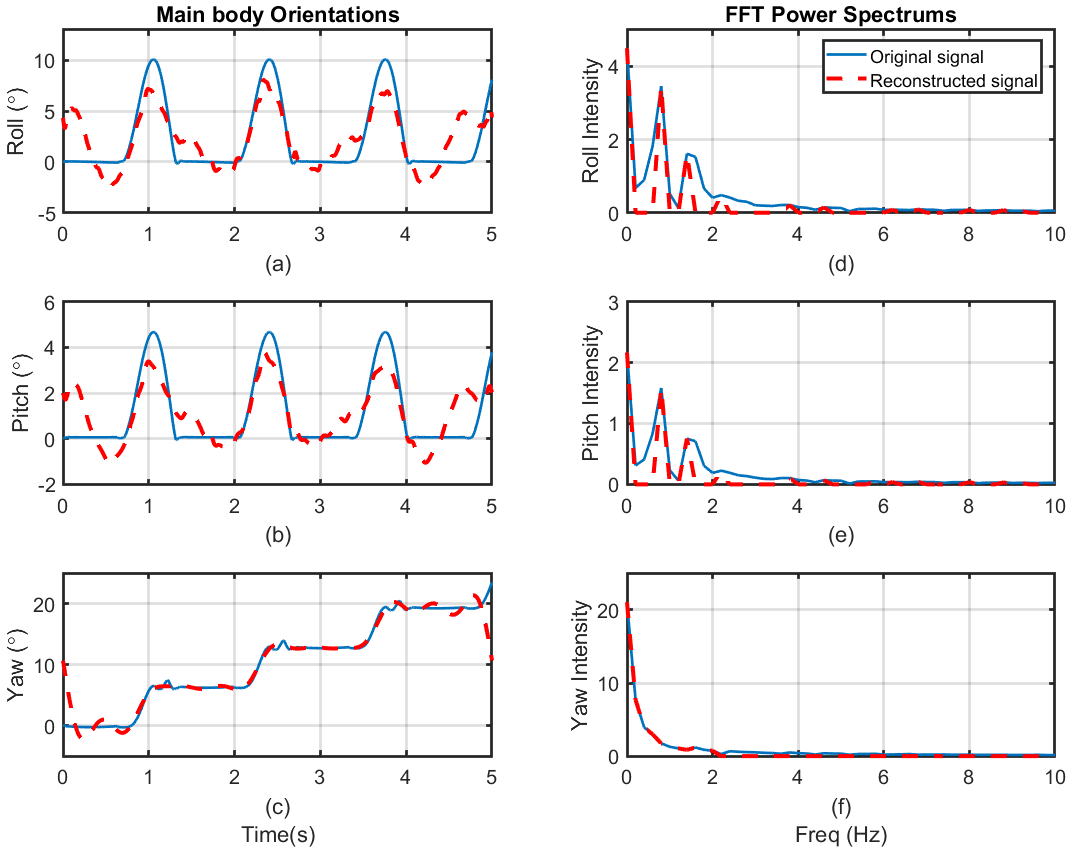}\\
  \caption{Robot main body orientation for identified morphology data: (a), (b), (c) Original and filtered signals in time domain; (d), (e), (f) Original and filtered PS plots in frequency domain}
  \label{fig:harsh1}
\end{figure}

\begin{figure}[hbt!]
  \centering
  \includegraphics[width=.35\textwidth]{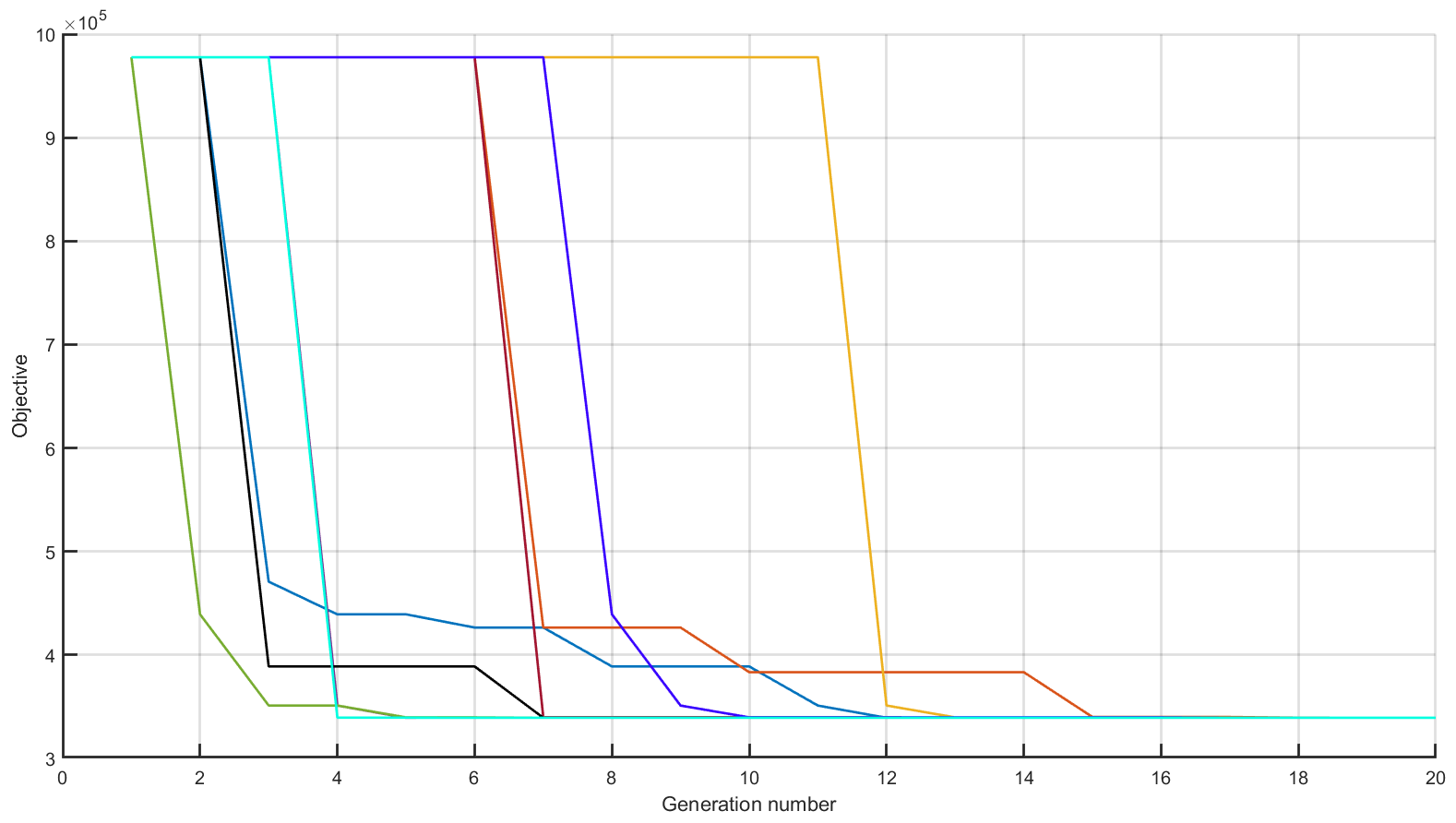}\\
    \caption{The best objective function value per generation for the harsh environment damage scenario}
  \label{fig:J_PLOT}
\end{figure}

To evaluate our damage identification algorithm under more realistic conditions, we created a challenging environment consisting of a 7-degree slope and sand and rocky surface, as depicted in Fig. \ref{fig:harsh}. The investigated damage scenario involved Legs 1 and 4 being nonfunctional. Despite this unmodeled environment, the FFT-based filter effectively processed the data, enabling the identification algorithm to detect the damage correctly, as shown in Figs. \ref{fig:harsh2} and \ref{fig:harsh1}. This demonstrates the robustness and efficiency of the proposed algorithm. Fig. \ref{fig:J_PLOT} shows the best objective function value per generation for this damage scenario. Across the 10 identification runs, the resulting morphology consistently indicated Leg 1’s first link is missing and Leg 4’s two last links are missing. While this outcome does not perfectly match the expected morphology of both legs being completely missing, it remains sufficiently accurate given the system's modeling limitations and the discrepancy discussed in the previous section. This result demonstrates the algorithm's ability to approximate the damaged structure, even in complex failure scenarios, providing a reliable foundation for the recovery process.


\section{Conclusion}\label{chap5:conclusion}

This paper presented a novel damage identification algorithm for MLRs, utilizing a fast, modular whole-body dynamical model. The algorithm efficiently detects physically damaged legs in under 10 minutes and demonstrates strong performance across various damage scenarios. By incorporating an FFT-based filter, the approach effectively mitigates the impact of communication delays, sensor noise, and disturbances, enhancing robustness in real-world applications. Testing in an unmodeled harsh environment confirmed the method’s reliability, as it maintained accurate identification despite unexpected body motions. While the algorithm performed well with limited sensory input, future improvements may include incorporating additional sensors, such as vision, to increase detection accuracy. Further research will also focus on identifying damage at the link level, optimizing robot trajectories for better diagnosis, and integrating the method into full damage recovery frameworks.

%In this paper, we presented a novel damage identification algorithm for MLR, leveraging a fast, modular, and whole-body dynamical model. The proposed algorithm effectively identifies physically damaged legs within a short time frame, averaging less than 10 minutes. Extensive evaluations across various damage scenarios demonstrated the accuracy of the method. Additionally, the FFT-based filtering approach proved highly effective in mitigating communication delays, sensory noise, and disturbances, enhancing the identification process's robustness.

%To validate the robustness of our approach, we tested the algorithm in an unmodeled harsh environment with a damage scenario. Despite the unexpected motions in the robot's main body, the FFT-based filter successfully reduced disturbances, enabling the algorithm to accurately identify damaged legs. While the damage detection algorithm faces challenges in harsh environments, incorporating additional sensory data, such as camera inputs, could further improve its accuracy.

%In future work, we aim to (i) enhance the algorithm to identify damage at the link level with greater precision, (ii) explore alternative trajectories for damaged robots to improve identification accuracy, and (iii) integrate the proposed method into damage recovery systems.

\bibliographystyle{IEEEtran}
    \bibliography{IEEEabrv,ref}

\end{document}